\documentclass{article}

\usepackage{microtype}
\usepackage{graphicx}
\usepackage{booktabs} 

\usepackage{hyperref}

\usepackage[accepted]{icml2025}

\usepackage{amsmath}
\usepackage{amssymb}
\usepackage{mathtools}
\usepackage{amsthm}
\usepackage{bm}                        
\usepackage{algorithm}
\usepackage{graphicx}
\usepackage{subcaption}
\usepackage{soul}
\usepackage{pifont}

\usepackage[capitalize,noabbrev]{cleveref}
\usepackage{multirow}

\theoremstyle{plain}

\theoremstyle{definition}

\theoremstyle{remark}

\usepackage[textsize=tiny]{todonotes}

\icmltitlerunning{Towards Building Non-Fine-Tunable Foundation Models}

\begin{document}

\twocolumn[
\icmltitle{Towards Building Non-Fine-Tunable Foundation Models}



\icmlsetsymbol{equal}{*}

\begin{icmlauthorlist}
\icmlauthor{Ziyao Wang}{yyy}
\icmlauthor{Nizhang Li}{comp}
\icmlauthor{Pingzhi Li}{sch}
\icmlauthor{Guoheng Sun}{yyy}
\icmlauthor{Tianlong Chen}{sch}
\icmlauthor{Ang Li}{yyy}
\end{icmlauthorlist}

\icmlaffiliation{yyy}{University of Maryland, College Park}
\icmlaffiliation{comp}{Macau University of Science and Technology}
\icmlaffiliation{sch}{The University
of North Carolina at Chapel Hill}

\icmlcorrespondingauthor{Ziyao Wang}{ziyaow@umd.edu}
\icmlcorrespondingauthor{Ang Li}{angliece@umd.edu}

\icmlkeywords{Machine Learning, ICML}

\vskip 0.3in
]



\printAffiliationsAndNotice{}  

\begin{abstract}
Open-sourcing foundation models (FMs) enables broad reuse but also exposes model trainers to economic and safety risks from unrestricted downstream fine-tuning.
We address this problem by building \emph{non-fine-tunable} foundation models: models that remain broadly usable in their released form while yielding limited adaptation gains under task-agnostic unauthorized fine-tuning.
We propose \textbf{Private Mask Pre-Training (PMP)}, a pre-training framework that concentrates representation learning into a sparse subnetwork identified early in training.
The binary mask defining this subnetwork is kept private, and only the final dense weights are released.
This forces unauthorized fine-tuning without access to the mask to update parameters misaligned with pretraining subspace, inducing an intrinsic mismatch between the fine-tuning objective and the pre-training geometry.
We provide theoretical analysis showing that this mismatch destabilizes gradient-based adaptation and bounds fine-tuning gains.
Empirical results on large language models demonstrating that PMP preserves base model performance while consistently degrading unauthorized fine-tuning across a wide range of downstream tasks, with the strength of non-fine-tunability controlled by the mask ratio.
\end{abstract}

\section{Introduction}

Foundation models (FMs), particularly large language models (LLMs), have achieved remarkable progress in recent years. Through large-scale pre-training on diverse data, these models exhibit strong general-purpose capabilities in language understanding, generation~\cite{achiam2023gpt,team2024gemini,comanici2025gemini}, and reasoning~\cite{guo2025deepseek,jaech2024openai}, and have become a core infrastructure for modern AI systems. Their impact now extends far beyond research, reshaping everyday applications such as information access, software development, education, and creative work. Training these models requires enormous investments in data, computation, and human expertise. In practice, FM trainers have followed two primary paths: keeping models closed and monetizing access through APIs~\cite{achiam2023gpt,comanici2025gemini}, or releasing pre-trained weights publicly~\cite{guo2025deepseek,touvron2023llama} to accelerate scientific innovation and community influence.

While open-weight releases have benefited the ecosystem by enabling advances in efficiency~\cite{lin2024awq}, interpretability~\cite{singh2024rethinking}, and post-training optimization~\cite{hu2022lora,yu2025dapo}, At the same time, they introduce two fundamental challenges for FM trainers. 
Economically, once a FM is released, users can perform low-cost adaptations on downstream tasks, effectively capturing value that would otherwise belong to the original trainer. 
From a safety perspective, trainers lose control over the model's downstream lifecycle: through unauthorized fine-tuning, an FM can be retrained on harmful or illegal data, such as content involving violence, extremism, discrimination, or other sensitive domains~\cite{li2026generative,poppi2024safe,wang2024moderator}. The resulting models may generate harmful outputs while still being associated with the original pre-trained FM, potentially damaging the trainer’s reputation and exposing them to legal or regulatory risk. 
\ul{\textit{These issues highlight the absence of mechanisms that allow trainers to open-source FMs while retaining meaningful control over adaptation.} }

Motivated by these concerns, we consider the problem of building \textbf{non-fine-tunable FMs}: pre-trained foundation models that remain broadly usable in their released form, but for which standard gradient-based fine-tuning without authorization yields substantially reduced gains.
To reconcile open-source utility with developer control, a non-fine-tunable FM must satisfy three core desiderata:
\ding{182} \textbf{Capability preservation}: the defense must not degrade the core performance, accuracy, or generation quality of the base model.
\ding{183} \textbf{Controllability}: unauthorized fine-tuning should be ineffective, while ``authorized" adaptation by trusted users remains possible.
\ding{184} \textbf{Black-box generality}: the FM trainer cannot assume any prior knowledge of the downstream user’s data, objectives, or optimization strategies.
Existing defenses often rely on task-specific unlearning  or ``immunization" strategies that solve a mini-max problem to suppress specific harmful domains~\cite{deng2024sophon, wang2024moderator}. These methods share a fundamental limitation: \textit{they require access to the data or task objectives they intend to restrict}. Such assumptions are unrealistic in real-world scenarios where trainers have no visibility into how a released model will be repurposed.

To address these challenges, we propose \textbf{Private Mask Pre-Training (PMP)}, a pre-training paradigm that induces non-fine-tunability as an intrinsic property of the model's pre-training geometry. PMP first runs a short warm-up pre-training phase on a small subset of data and uses an early-bird lottery ticket criterion~\cite{chen2021earlybert} to select a sparse binary mask that identifies a stable subnetwork. The mask is then kept private and applied throughout pre-training so that all gradient updates are restricted to the masked parameters, while the masked-out parameters remains frozen. Although the forward pass uses the full dense parameterization, representation learning is concentrated into the selected sparse subnetwork. After convergence, the model trainer releases only the final dense weights and does not disclose the mask. Without access to the mask, unauthorized fine-tuning cannot restrict updates to the true subnetwork and is forced to co-update both the trained ticket and the untrained complement, creating an intrinsic mismatch between the conditional objective solved during pre-training and the joint objective optimized during fine-tuning. This mismatch destabilizes optimization and significantly degrades adaptation, as formalized by our theoretical analysis. Importantly, our experiments demonstrate that with a reasonable mask ratio (e.g., 70\%), PMP preserves the pre-trained performance of the base model, while substantially degrading unauthorized fine-tuning across diverse downstream tasks, datasets, and optimization settings, without requiring any prior knowledge of downstream data.
When the private mask is provided, authorized fine-tuning can partially mitigate the degradation caused by PMP, often achieving better performance than unauthorized fine-tuning.


We summarize our main contributions as follows:
\begin{itemize}
    \item We identify non-fine-tunability as a fundamental pre-training-level problem for FMs and articulate the key challenges faced by model trainers in controlling unauthorized fine-tuning.
    \item We propose PMP, a general pre-training paradigm that preserves base model utility, enables authorized fine-tuning, and restricts unauthorized fine-tuning in a task-agnostic black-box setting.
    \item We provide a theoretical analysis demonstrating that PMP induces an intrinsic optimization objective mismatch that injects destabilizing perturbations during unauthorized updates.
    \item We empirically validate PMP on LLMs, showing that it consistently suppresses adaptation gains across diverse benchmarks while maintaining competitive pre-trained performance.   
\end{itemize}

\begin{figure*}[t]
  \centering
  \includegraphics[width=0.9\textwidth]{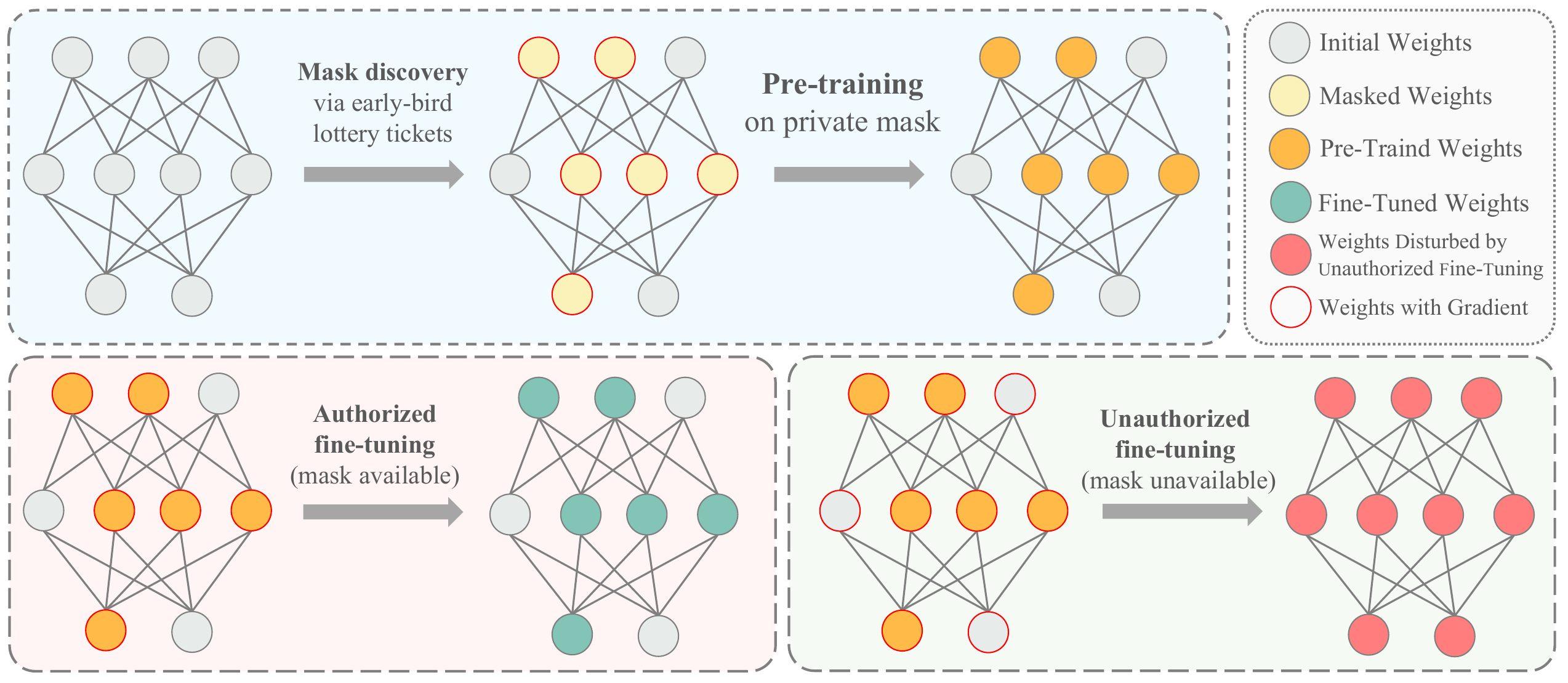}
  \caption{Overview of Private Mask Pretraining (PMP).
Top: the foundation model trainer discovers an Early-Bird lottery ticket and performs pretraining by restricting updates to a private mask.
Bottom: after model release, authorized fine-tuning with access to the mask updates only the pretrained ticket and enables effective adaptation, while unauthorized fine-tuning without the mask perturbs both trained and untrained parameters, leading to degraded performance.}

  \label{fig:pmp-overview}
\end{figure*}

\section{Method}
\label{sec:method}

We propose PMP to obtain a non-fine-tunable FM. As shown in Figure~\ref{fig:pmp-overview}, the core idea is to concentrate pre-training into a sparse subnetwork identified early in training, while keeping the corresponding mask private. After pre-training, only the final dense parameters are released. Without access to the mask, downstream fine-tuning is forced to update the full parameter vector in a way that is misaligned with how the model was pre-trained, which makes optimization unstable and can substantially deteriorate fine-tuning outcomes.

PMP proceeds in three stages: (I) private early-bird mask discovery that identifies a sparse ticket; (II) private mask pre-training that updates only the ticket parameters; and (III) releasing final weights without the mask.

\subsection{Private Early-Bird Mask Discovery}

Let model parameters be $\theta \in \mathbb{R}^d$ with initialization $\theta^{(0)}$. We seek a binary mask $M \in \{0,1\}^d$ at sparsity ratio $\rho \in (0,1)$; denote its complement by $\bar M = \mathbf{1}-M$. Masked parameters are defined via element-wise product $\odot$:
\begin{equation}
\label{eq:mask_defs}
\theta_M = \theta \odot M, \qquad
\theta_{\bar M} = \theta \odot \bar M, \qquad
\|M\|_0 = \rho d .
\end{equation}

Using a small warm-up subset $\mathcal{D}_{\mathrm{pre}}$, we identify an early-bird lottery ticket by extracting a stable global top-$\rho$ mask from gradient magnitudes. In the warm-up step $t$, let $\mathcal{B}^{(t)} \subset \mathcal{D}_{\mathrm{pre}}$ denote the mini-batch sampled at step $t$, and let
\[
g^{(t)} = \big|\nabla_{\theta}\mathcal{L}(\theta^{(t)};\mathcal{B}^{(t)})\big|\in\mathbb{R}^d_{\ge 0}
\]
be the element-wise absolute gradient. We form a candidate mask $M^{(t)}\in\{0,1\}^d$ by keeping exactly the top-$k$ entries of $g^{(t)}$ globally across all parameters, where $k=\lfloor \rho d\rfloor$:
\begin{equation}
\label{eq:topk_mask}
M^{(t)}_i=\mathbf{1}\!\left[g^{(t)}_i \ge \mathrm{TopK}\!\big(g^{(t)},k\big)\right],\qquad i=1,\dots,d,
\end{equation}
where $\mathrm{TopK}(g,k)$ returns the $k$-th largest value of $g$ (ties broken arbitrarily to ensure $\|M^{(t)}\|_0=k$). We declare the early-bird ticket once consecutive masks become stable, measured by the IoU
\begin{equation}
\label{eq:iou}
\mathrm{IoU}(M^{(t)},M^{(t-1)})=\frac{\|M^{(t)}\odot M^{(t-1)}\|_0}{\|M^{(t)}\vee M^{(t-1)}\|_0},
\end{equation}
and select the first $M^{(t)}$ that remains above a threshold for several successive steps. The resulting mask $M$ is then fixed and kept private for all subsequent training:
\begin{equation}
\label{eq:warmup}
M \leftarrow \mathrm{EarlyBird}\!\big(\{\nabla \mathcal{L}(\theta^{(t)};\mathcal{B}^{(t)})\}_{t \le t_{\mathrm{EB}}},\, \rho\big).
\end{equation}

\subsection{Private Mask Pre-Training}

Let $\mathcal{D}_{\mathrm{train}}$ be the full pre-training dataset. Starting from $\theta^{(0)}$, PMP performs standard pre-training while restricting optimization to the ticket parameters $\theta_M$. At each iteration, a mini-batch $\mathcal{B} \subset \mathcal{D}_{\mathrm{train}}$ is sampled, and gradient descent is applied only to $\theta_M$, while the complement $\theta_{\bar M}$ is kept fixed:
\begin{equation}
\label{eq:ticket_step}
\theta_M \leftarrow \theta_M - \eta \nabla_{\theta_M}
\mathcal{L}\big(\theta_M + \theta_{\bar M}; \mathcal{B}\big),
\quad
\theta_{\bar M} \leftarrow \theta_{\bar M}.
\end{equation}

Although the forward computation uses the full parameter vector $\theta = \theta_M + \theta_{\bar M}$, all parameter updates are confined to the sparse ticket defined by $M$. This concentrates learned representations into a low-dimensional subspace determined early in training, while leaving the complement parameters unadapted.


\subsection{Release without Mask}

At convergence, we publish only the final parameters
and keep the mask $M$ private. Since downstream users do not know which parameters constitute the true ticket, fine-tuning necessarily updates the full parameter vector. This creates a mismatch between the constrained optimization used during pre-training (updates restricted to $\theta_M$) and the unconstrained optimization used during adaptation (updates applied to both $\theta_M$ and $\theta_{\bar M}$), which can destabilize optimization and lead to substantially worse fine-tuning results. We next provide a theoretical explanation of this mismatch effect.

\section{Theoretical Analysis}
\label{sec:theory}

We provide a theoretical explanation for why fine-tuning a PMP-pre-trained model without access to the private mask can become unstable and yield only limited adaptation gains. Our analysis formalizes \emph{non-fine-tunability} as a property of the achievable fine-tuning improvement under unknown-mask optimization, and explains how PMP enforces this property through an intrinsic optimization mismatch.

We decompose parameters as $\theta=(\theta_M,\theta_{\bar M})$ using the mask in Eq.~\eqref{eq:mask_defs}, and define the expected pre-training objective
\begin{equation}
L_{\mathrm{pre}}(\theta_M,\theta_{\bar M})
\;=\;
\mathbb{E}_{z\sim \mathcal{D}_{\mathrm{train}}}
\big[\ell(f(\theta_M+\theta_{\bar M});z)\big].
\end{equation}
Let $(\theta_M^\star,\theta_{\bar M}^{(0)})$ denote the point produced by PMP, where $\theta_{\bar M}^{(0)}$ is the frozen complement throughout pre-training. We assume that PMP reaches a \emph{conditional optimum}, where $\theta_M$ is optimal given fixed $\theta_{\bar M}^{(0)}$, but not necessarily optimal over the full parameter space:
\[
\theta_M^\star \in \arg\min_{\theta_M} L_{\mathrm{pre}}(\theta_M,\theta_{\bar M}^{(0)}),
\quad
\nabla_{\theta_M} L_{\mathrm{pre}}(\theta_M^\star,\theta_{\bar M}^{(0)}) = 0.
\]

\paragraph{Fine-tuning gain and non-fine-tunability.}
For a downstream task, define the fine-tuning objective as
\begin{equation}
L_{\mathrm{ft}}(\theta_M,\theta_{\bar M})
\;=\;
L_{\mathrm{pre}}(\theta_M,\theta_{\bar M})
\;+\;
\Delta(\theta_M,\theta_{\bar M}),
\end{equation}
where $\Delta$ captures task-specific supervision or distributional shift.
We measure fine-tuning effectiveness by the improvement in pre-training loss,
$
\mathrm{Gain} = L_{\mathrm{pre}}(\theta_M^\star,\theta_{\bar M}^{(0)})
-
L_{\mathrm{pre}}(\theta_M,\theta_{\bar M}),
$
achieved through optimization on $L_{\mathrm{ft}}$.
A model is \emph{non-fine-tunable} if this gain remains small under unknown-mask fine-tuning.

\textbf{Assumption 1 (Asymmetric local geometry).}
In a neighborhood of $(\theta_M^\star,\theta_{\bar M}^{(0)})$, the loss landscape of $L_{\mathrm{pre}}$ is comparatively flat along the $\theta_M$ directions due to extensive optimization, while exhibiting non-negligible positive curvature along the $\theta_{\bar M}$ directions, which were never adapted during PMP pre-training.

\textbf{Proposition 1 (Instability and limited gains under unknown-mask fine-tuning).}
Consider one step of stochastic gradient descent on $L_{\mathrm{ft}}$ starting from
$(\theta_M^\star,\theta_{\bar M}^{(0)})$ with step size $\eta$:
\[
\theta_M^{+} = \theta_M^\star - \eta g_M,
\qquad
\theta_{\bar M}^{+} = \theta_{\bar M}^{(0)} - \eta g_{\bar M},
\]
where $(g_M,g_{\bar M})$ is a stochastic gradient of $L_{\mathrm{ft}}$ at
$(\theta_M^\star,\theta_{\bar M}^{(0)})$.
Under Assumption~1 and for sufficiently small $\eta$,
\begin{equation}
\mathbb{E}\!\left[
L_{\mathrm{pre}}(\theta_M^{+},\theta_{\bar M}^{+})
\right]
\;\ge\;
L_{\mathrm{pre}}(\theta_M^\star,\theta_{\bar M}^{(0)})
\;+\;
c\,\eta^2,
\end{equation}
for some constant $c>0$ whenever $g_{\bar M}\neq 0$ with nonzero probability.
As a result, when fine-tuning is performed without the private mask and induces nontrivial updates on the complement parameters $\theta_{\bar M}$, the optimization dynamics are \emph{prone to destabilizing perturbations} of the pre-trained solution.
Consequently, unknown-mask fine-tuning \emph{often yields limited or unstable fine-tuning gains}, as progress driven by task-specific gradients is counteracted by curvature-induced degradation along previously unadapted directions.

\textbf{Proof sketch.}
We perform a second-order Taylor expansion of $L_{\mathrm{pre}}$ around
$(\theta_M^\star,\theta_{\bar M}^{(0)})$:
{\small
\[
L_{\mathrm{pre}}(\theta_M^{+},\theta_{\bar M}^{+})
\approx
L_{\mathrm{pre}}(\theta_M^\star,\theta_{\bar M}^{(0)})
-\eta \langle \nabla L_{\mathrm{pre}}, g \rangle
+\frac{\eta^2}{2} g^\top H_{\mathrm{pre}} g,
\]
}
where $g=(g_M,g_{\bar M})$ and $H_{\mathrm{pre}}$ is the Hessian of
$L_{\mathrm{pre}}$ at $(\theta_M^\star,\theta_{\bar M}^{(0)})$.
Since
$\nabla_{\theta_M} L_{\mathrm{pre}}(\theta_M^\star,\theta_{\bar M}^{(0)})=0$,
the first-order term depends only on the complement component
$g_{\bar M}$ and has no guaranteed sign.
Under Assumption~1, the Hessian block corresponding to $\theta_{\bar M}$
exhibits positive curvature, so the quadratic term contributes positively
in expectation whenever $g_{\bar M}\neq 0$.
This explains why unconstrained fine-tuning that co-updates $\theta_{\bar M}$
tends to destabilize the pre-trained solution and limits effective
adaptation without access to the private mask.

\begin{table*}[t]
\centering
\setlength{\tabcolsep}{5pt}
\renewcommand{\arraystretch}{1.15}
\caption{GLUE \textit{dev} results (\%). For each model (GPT-2, TinyLlama) and each stage (Base, LoRA fine-tuned, Full fine-tuned), we report results \textit{w/o} PMP and with \textbf{Private Mask Pre-Training (PMP)} under identical pre-training and fine-tuning protocols. Base (↑): higher is better; Fine-Tuned (↓): lower is better.}

\label{tab:glue_pmp_rows}
\resizebox{\textwidth}{!}{%
\begin{tabular}{l | c | c | c c c c c c c c c}
\toprule
Model & Stage & Variant
& CoLA & SST-2 & MRPC & QQP & STS-B & MNLI-m & MNLI-mm & QNLI & RTE \\
\midrule
\multirow{6}{*}{GPT-2} & \multirow{2}{*}{Base($\uparrow$)} & W/O PMP
& 38.35  & 50.00  & 67.16  & 37.22  &12.52  & 35.90  & 36.02  & 49.35  & 52.71 \\
&  & PMP & 50.34  & 50.00  & 58.63  &37.42  &11.88  & 35.00  & 35.10  & 48.20  & 51.00  \\
\cline{2-12}
& \multirow{2}{*}{LoRA Fine-Tuned($\downarrow$)} & W/O PMP &57.53 &68.58 &68.38 &78.42 &17.40 &40.89 &45.05 &51.58 &57.76 \\
&  & PMP
& \textbf{52.73} & \textbf{57.91} & \textbf{56.13} &\textbf{71.68}  &\textbf{14.67} & \textbf{32.72} & \textbf{42.34} & \textbf{45.25} & \textbf{49.82} \\
\cline{2-12}
& \multirow{2}{*}{Full Fine-Tuned($\downarrow$)} & W/O PMP & 53.49  & 79.24  & 68.14  &56.61  &20.58  & 36.41  & 39.05  & 51.24  & 55.60 \\
&  & PMP
& \textbf{50.05}  & \textbf{65.94}  & \textbf{61.40}  &\textbf{38.81}  &\textbf{18.30}  & \textbf{33.28}  & \textbf{31.80}  & \textbf{50.54}  & \textbf{47.29}  \\

\midrule
\multirow{6}{*}{TinyLlama}
& \multirow{2}{*}{Base($\uparrow$)} & W/O PMP
&37.01  &50.69  &31.62  &36.84  & 16.13 &36.77  & 33.35 & 50.58 & 52.71  \\
&  & PMP
&63.66  &50.80 &40.20  &38.66  &16.53  &32.81  &32.60  &50.72  &53.79  \\
\cline{2-12}
& \multirow{2}{*}{LoRA Fine-Tuned($\downarrow$)} & W/O PMP
&68.36  &80.62  &71.08  &82.10  &26.33 &66.91  &68.65  & 78.73 &54.51 \\
&  & PMP
& \textbf{65.96} &\textbf{72.13} &\textbf{63.97}  &\textbf{74.66}  & \textbf{22.40} &\textbf{61.15}  &\textbf{62.42}  &\textbf{62.47} &\textbf{44.04} \\
\cline{2-12}
& \multirow{2}{*}{Full Fine-Tuned($\downarrow$)} & W/O PMP
&68.36  &83.14  &71.32  &86.51  &32.47  &71.17  &71.34 &79.37  & 54.15  \\
&  & PMP
&\textbf{61.64}  &\textbf{76.03}  &\textbf{69.36}  &\textbf{78.20} & \textbf{22.43} &\textbf{59.52} & \textbf{58.48} &\textbf{63.75} &\textbf{46.57}  \\
\bottomrule
\end{tabular}}
\vspace{-0.5mm}
\footnotesize \\
\end{table*}

\section{Experiments}
\subsection{Experimental Set-Ups}

\paragraph{Model, data, and initialization.}
We pre-train a LLaMA-style decoder-only Transformer with a TinyLlama-1.1B--scale configuration (22 layers, hidden size 2048, 32 attention heads, 4 KV heads, intermediate size 5632, RoPE $\theta=10{,}000$) from random initialization.
Pre-training is performed on SlimPajama-6B (\texttt{DKYoon/SlimPajama-6B}, \texttt{text} field) using the HuggingFace streaming pipeline with online shuffling (buffer size 50k).
Documents are tokenized without truncation, separated by an EOS token, and packed into fixed-length blocks of 256 tokens.
Each block is trained with the standard causal language modeling objective, with labels equal to input IDs.
Both \textsc{Base} and \textsc{PMP} models use the same architecture and training budget.

\paragraph{\textsc{PMP} procedure.}
We run $t_{\mathrm{EB}}=500$ warm-up iterations and construct a global exact Top-$k$ mask from element-wise absolute gradients, with sparsity $\rho=0.70$. We select the earliest mask whose IoU is at least $0.99$ for 5 steps.

\begin{figure*}[t]
  \centering
  \begin{minipage}{0.9\textwidth}
  \begin{subfigure}[t]{0.48\linewidth}
    \centering
    \includegraphics[width=\linewidth]{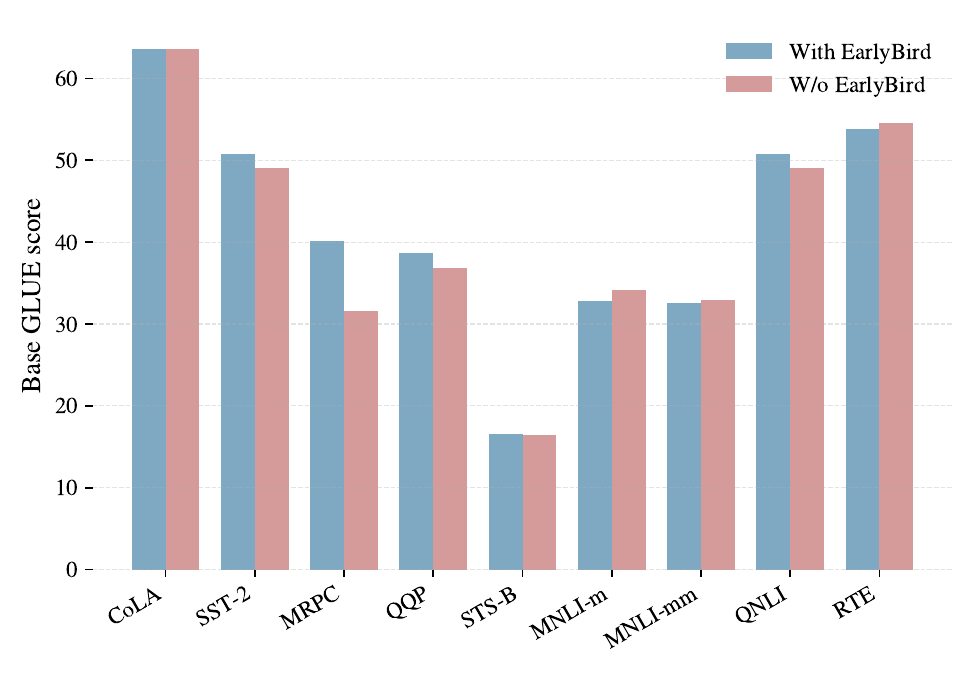}
    \caption{Base Acc. Higher is better.}
    \label{fig:earlybird-base}
  \end{subfigure}\hfill
  \begin{subfigure}[t]{0.48\linewidth}
    \centering
    \includegraphics[width=\linewidth]{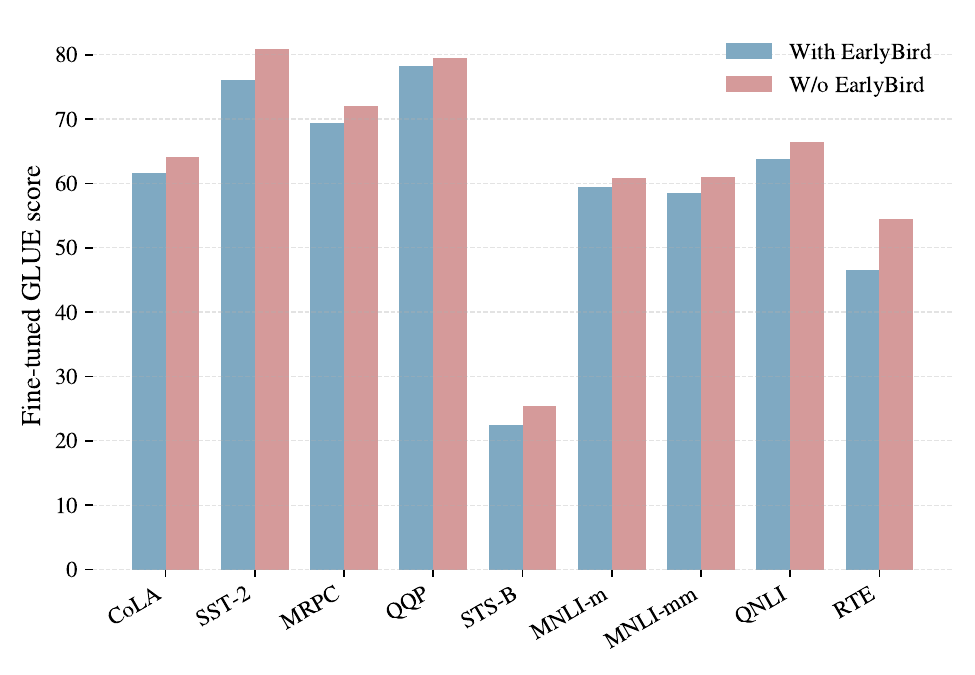}
    \caption{Fine-tuned Acc. Lower is better.}
    \label{fig:earlybird-ft}
  \end{subfigure}
  \end{minipage}
  \caption{\textbf{Effect of EarlyBird mask selection on TinyLlama in PMP.} We compare PMP with EarlyBird and PMP with random mask selection on (a) base performance and (b) fine-tuned performance.}
  \label{fig:earlybird-ablation}
\end{figure*}

\paragraph{Downstream fine-tuning.}
We evaluate fine-tunability on GLUE~\cite{wang2018glue} by fine-tuning each released checkpoint independently on the official training split and reporting results on the validation split.
For tasks whose primary metric is accuracy (SST-2, MNLI matched/mismatched, QNLI, and RTE), we follow the standard GLUE protocol and report accuracy; for pairwise sentence classification tasks (MRPC and QQP), we report the standard GLUE metrics (accuracy and F1).
However, some GLUE tasks adopt canonical metrics that are not on the same scale as accuracy (e.g., MCC for CoLA and Pearson/Spearman correlation for STS-B), making direct cross-task aggregation (e.g., an average score) ill-posed.
To enable a unified, accuracy-based aggregate across tasks, we additionally introduce an \emph{accuracy-style auxiliary metric} for CoLA and STS-B by mapping their outputs to a form amenable to accuracy computation.
This auxiliary metric is used solely to improve cross-task comparability and aggregation, and is not intended to replace the canonical task metrics.
Specifically, for \textbf{CoLA} (binary grammatical acceptability), we take the argmax over the two logits to obtain the predicted label $\hat{y}$ and compute accuracy as the fraction of examples with $\hat{y}=y$.
For \textbf{STS-B} (semantic textual similarity; regression with gold scores in $[0,5]$), we discretize both predictions and gold scores by clipping to $[0,5]$ and rounding to the nearest integer, yielding ordinal labels in $\{0,1,2,3,4,5\}$, and then compute standard classification accuracy on these discretized labels.

\subsection{The Non-Fine-Tunability of PMP}

We evaluate the non-fine-tunability of PMP on the GLUE development benchmarks, comparing models trained with and without PMP under identical pretraining and fine-tuning protocols. Table~1 reports results for GPT-2 and TinyLlama, where we separately examine base performance before fine-tuning and performance after standard fine-tuning without access to the private mask. For base models, higher values indicate better performance, while for fine-tuned models, lower values indicate worse adaptation.

Across both architectures, PMP preserves the pretrained capability of the released base models. For GPT-2, PMP achieves base performance comparable to training without PMP across all GLUE tasks, with only minor fluctuations that do not indicate systematic degradation. Similarly, for TinyLlama, PMP maintains strong base performance and even improves results on several tasks such as CoLA and MRPC, demonstrating that concentrating representation learning into a sparse ticket during pretraining does not harm general-purpose capability.
In contrast, fine-tuning behavior differs markedly. When fine-tuning is performed without access to the private mask, models pretrained with PMP consistently exhibit substantially worse fine-tuned performance than their non-PMP counterparts. For GPT-2, PMP leads to clear degradation across nearly all tasks, including SST-2, MRPC, QQP, MNLI, QNLI, and RTE. The same pattern is observed for TinyLlama, where PMP results in significantly lower fine-tuned scores across the full suite of GLUE benchmarks. Notably, this degradation is broad and consistent rather than confined to a small subset of tasks, indicating a general failure of effective adaptation rather than task-specific brittleness.

Overall, these results demonstrate a clear decoupling between pretrained capability and fine-tunability. While PMP-pretrained models remain competitive foundation models in their released form, unauthorized fine-tuning without the private mask becomes substantially less effective across architectures and tasks. This empirical evidence supports PMP as a practical pretraining-level mechanism for inducing non-fine-tunability without sacrificing base model quality.

\subsection{Authorized Fine-Tuning Performance}
\label{sec:auth_ft}

A central goal of PMP is to preserve practical usability while enabling \emph{authorized} adaptation. In this setting, the model owner provides the private mask to the user, who then fine-tunes by updating only the masked subspace (i.e., $\theta_M$) while keeping the complement fixed. We compare this authorized procedure against \emph{unauthorized} fine-tuning, which applies standard full-parameter updates without the private mask.

\begin{figure}[t]
  \centering
  \includegraphics[width=0.9\linewidth]{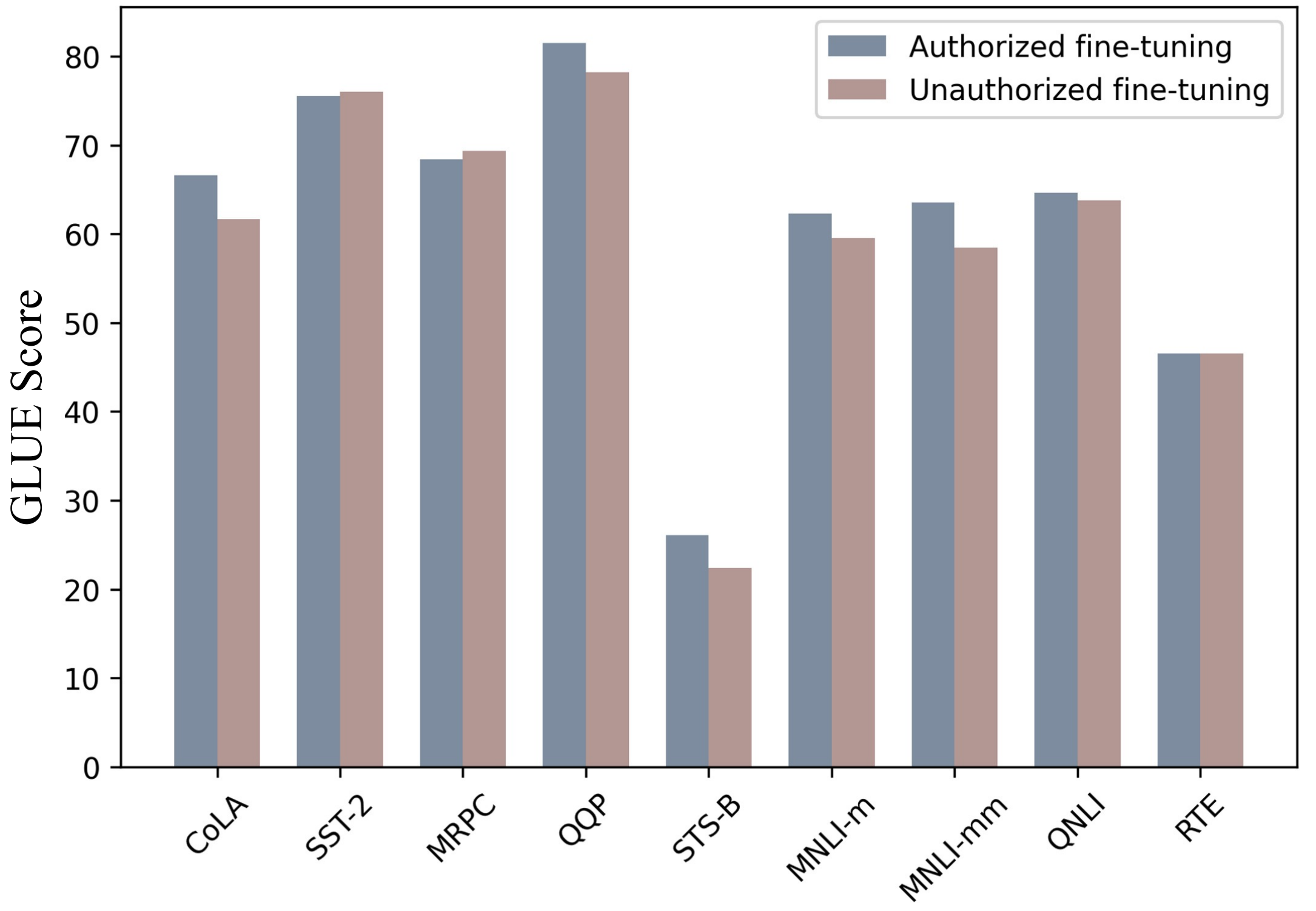}
  \caption{\textbf{Authorized vs.\ unauthorized fine-tuning on GLUE (TinyLlama).}
  Authorized fine-tuning (with access to the private mask) updates only the masked subspace, while unauthorized fine-tuning performs standard full-parameter updates without the mask. Authorized fine-tuning consistently outperforms unauthorized fine-tuning.}
  \label{fig:auth_vs_unauth_glue}
\end{figure}

Figure~\ref{fig:auth_vs_unauth_glue} shows the resulting GLUE performance on TinyLlama. Authorized fine-tuning consistently matches or exceeds unauthorized fine-tuning across tasks, with the largest gaps appearing on CoLA, QQP, STS-B, and MNLI. This trend supports our core hypothesis: when the fine-tuning updates are aligned with the pre-trained ticket (authorized access), adaptation is more stable and effective, whereas unconstrained full-space updates without the mask tend to underperform due to the induced optimization mismatch.

\subsection{Ablation Studies}
\paragraph{Early-Bird Mask Selection.}
We further analyze the effect of EarlyBird mask selection by explicitly studying its role as an early-bird lottery ticket in PMP. In this experiment, we compare PMP using the EarlyBird lottery ticket mask discovered during warm-up with a variant that uses a randomly sampled mask of the same sparsity ratio $\rho$. All other components, including the model architecture, pretraining schedule, and unauthorized fine-tuning protocol, are kept identical. As shown in Figure~\ref{fig:earlybird-ablation}, both variants maintain comparable base performance, confirming that sparsity alone does not inherently degrade pretrained capability. Notably, PMP with the EarlyBird lottery ticket mask consistently achieves equal or slightly better base performance across most GLUE tasks, indicating that the early-bird ticket captures a subnetwork that is particularly effective for representation learning during pretraining.

The impact of EarlyBird selection is substantially more pronounced under fine-tuning. When the EarlyBird lottery ticket mask is replaced by a random mask, fine-tuned performance improves across most tasks, suggesting that the model becomes easier to adapt under unauthorized fine-tuning. In contrast, PMP with the EarlyBird lottery ticket mask yields consistently lower fine-tuned scores, demonstrating a stronger non-fine-tunability effect. This behavior aligns with the lottery ticket hypothesis: the EarlyBird mask identifies a sparse subnetwork that emerges early and remains stable throughout training, concentrating pretrained representations into a well-organized parameter subspace. As a consequence, pretraining becomes more effective within this ticket, while unauthorized fine-tuning, which cannot target the same subspace, is more likely to disrupt the learned structure and result in degraded adaptation. Together, these results show that using an EarlyBird lottery ticket mask is critical for simultaneously improving pretraining effectiveness and amplifying the non-fine-tunability induced by PMP.

\begin{figure*}[t]
  \centering
  \includegraphics[width=0.99\textwidth]{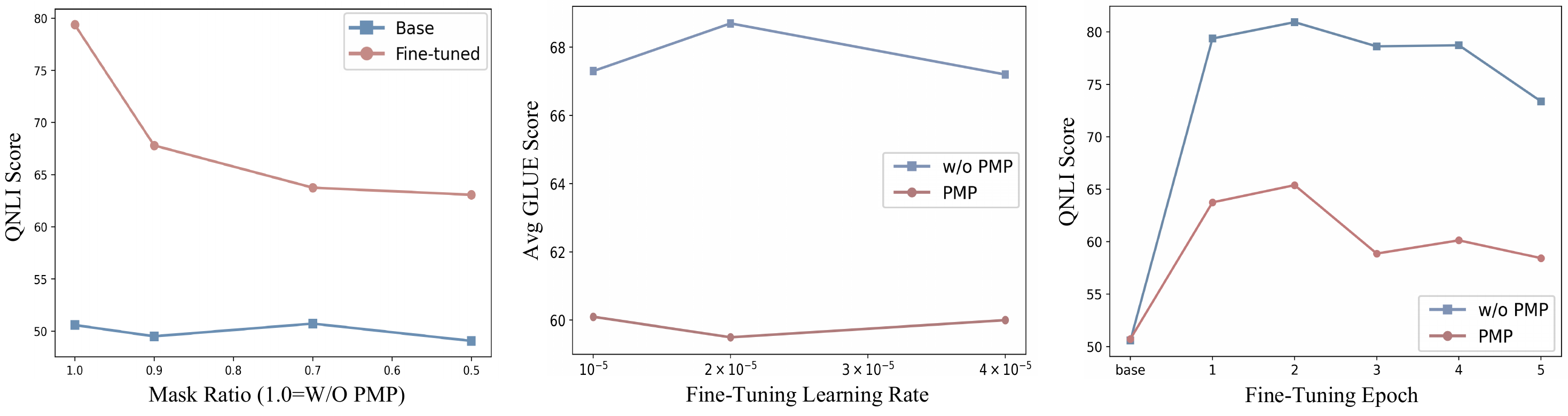}
  \caption{\textbf{Impact of Mask Ratio and Optimization Hyperparameters on Fine-Tuning Performance.}
The left panel shows QNLI performance under different mask ratios, where standard fine-tuning yields substantial gains while PMP consistently suppresses adaptation with minimal impact on base performance. The middle panel evaluates sensitivity to the fine-tuning learning rate, and the right panel reports performance across fine-tuning epochs. Across all settings, PMP consistently limits downstream adaptation, and increasing optimization strength does not recover performance.}

  \label{fig:ablation}
\end{figure*}

\paragraph{Mask Ratio.}
The mask ratio determines the fraction of parameters included in the pretrained ticket and directly controls the strength of non-fine-tunability.
Figure~\ref{fig:ablation} (left) shows QNLI base and fine-tuned accuracy as a function of the mask ratio, where $1.0$ corresponds to training without PMP.
Without PMP, fine-tuning yields substantial gains, indicating strong adaptation capacity.
In contrast, once PMP is applied, fine-tuning improvements are sharply suppressed and remain limited across mask ratios from $0.9$ to $0.5$, while base performance remains largely stable.
Among these settings, a mask ratio of $0.7$ achieves the best trade-off between base accuracy and fine-tuning suppression, and is therefore used as the default throughout our experiments.
Overall, these results demonstrate that the mask ratio serves as a continuous control knob that modulates the degree of fine-tuning suppression induced by PMP.



\paragraph{Fine-Tuning Learning Rate.}
We examine whether the degradation induced by PMP can be mitigated by tuning the fine-tuning learning rate. Figure~\ref{fig:ablation} (middle) reports the average GLUE performance under three commonly used learning rates, comparing standard fine-tuning without PMP and fine-tuning on PMP models. Without PMP, fine-tuning performance remains consistently high across learning rates, indicating that standard pretraining supports robust adaptation under typical hyperparameter choices. In contrast, PMP consistently suppresses fine-tuning performance for all learning rates considered. The performance gap between PMP and the baseline remains largely unchanged as the learning rate varies, and increasing the learning rate does not recover fine-tuning effectiveness. This shows that the non-fine-tunability induced by PMP is not a consequence of suboptimal hyperparameter selection, but rather reflects a fundamental mismatch between the pretraining and fine-tuning optimization objectives. Consequently, common learning-rate tuning strategies are insufficient to bypass PMP.

\paragraph{Fine-Tuning Epoch.}
We further examine the effect of fine-tuning duration by varying the number of training epochs. As shown in Figure~\ref{fig:ablation} (right), standard fine-tuning rapidly improves downstream performance but degrades with over-training, whereas PMP consistently suppresses adaptation across epochs. Notably, increasing the number of fine-tuning epochs does not recover performance under PMP, indicating that its effect is not sensitive to training duration.

\section{Discussion and Analysis}
\label{sec:discussion}

\subsection{Understanding PMP via Loss Landscape}
To empirically validate the ``Optimization Mismatch'' hypothesis proposed in Section~\ref{sec:theory}, we visualize the loss landscape geometry of the PMP-trained model. Unlike standard pre-training, which typically seeks a minimum that is flat in all directions to improve generalization, PMP theoretically creates a highly \textit{anisotropic} landscape—flat within the authorized subspace but sharp along the frozen directions.

\begin{figure}[h]
    \centering
    \includegraphics[width=0.98\linewidth]{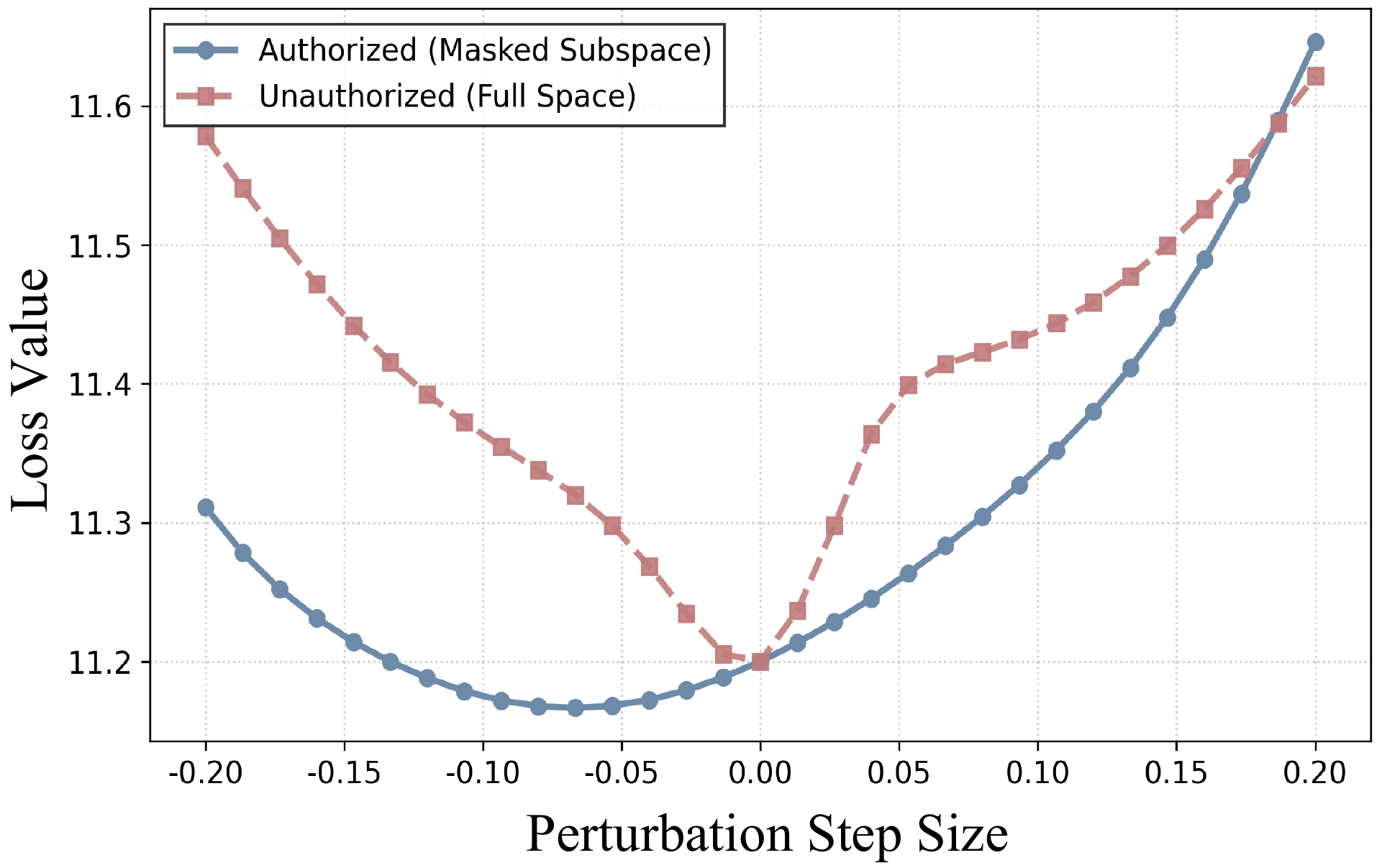}
    \vspace{2pt}
    \caption{\textbf{1D Loss Landscape Visualization.} We compare the loss curve along an authorized direction (confined to the mask, \textcolor[HTML]{6D8AA8}{\textbf{Blue}}) versus an unauthorized direction (full parameter space, \textcolor[HTML]{C07B7B}{\textbf{Red}}).}
    \label{fig:landscape}
\end{figure}

We verify this by plotting the 1D loss interpolation $f(\alpha) = \mathcal{L}(\theta^* + \alpha \cdot \delta)$ along two distinct random directions: (1) \textbf{Authorized Direction ($\delta_M$):} A random vector normalized and confined strictly within the private mask subspace (i.e., $\delta_M \odot (1-M) = 0$). (2) \textbf{Unauthorized Direction ($\delta_{Full}$):} A random vector normalized in the full parameter space, representing the trajectory of standard, unauthorized fine-tuning.
As illustrated in Figure~\ref{fig:landscape}, the geometric distinction is pronounced. The \textbf{Authorized (Blue)} trajectory remains within a low-loss basin, exhibiting a smooth and flat curvature that preserves the model's pre-trained capabilities. In sharp contrast, the \textbf{Unauthorized (Red)} trajectory immediately encounters \textbf{steep optimization barriers}. Even with minute perturbations ($\alpha \approx \pm 0.1$), the loss along the full space direction spikes significantly. 

This visualization provides a physical intuition for how PMP works: without applying the private mask, gradient updates inevitably possess components along the frozen, high-curvature directions (the ``steep walls'' of the valley). This forces the model parameters to ascend these sharp barriers, leading to the rapid destabilization and performance degradation observed in our main experiments. The mismatched objectives between pre-training and fine-tuning form a pitfall for unauthorized fine-tuning.

\begin{figure}[t]
    \centering
    \includegraphics[width=0.98\linewidth]{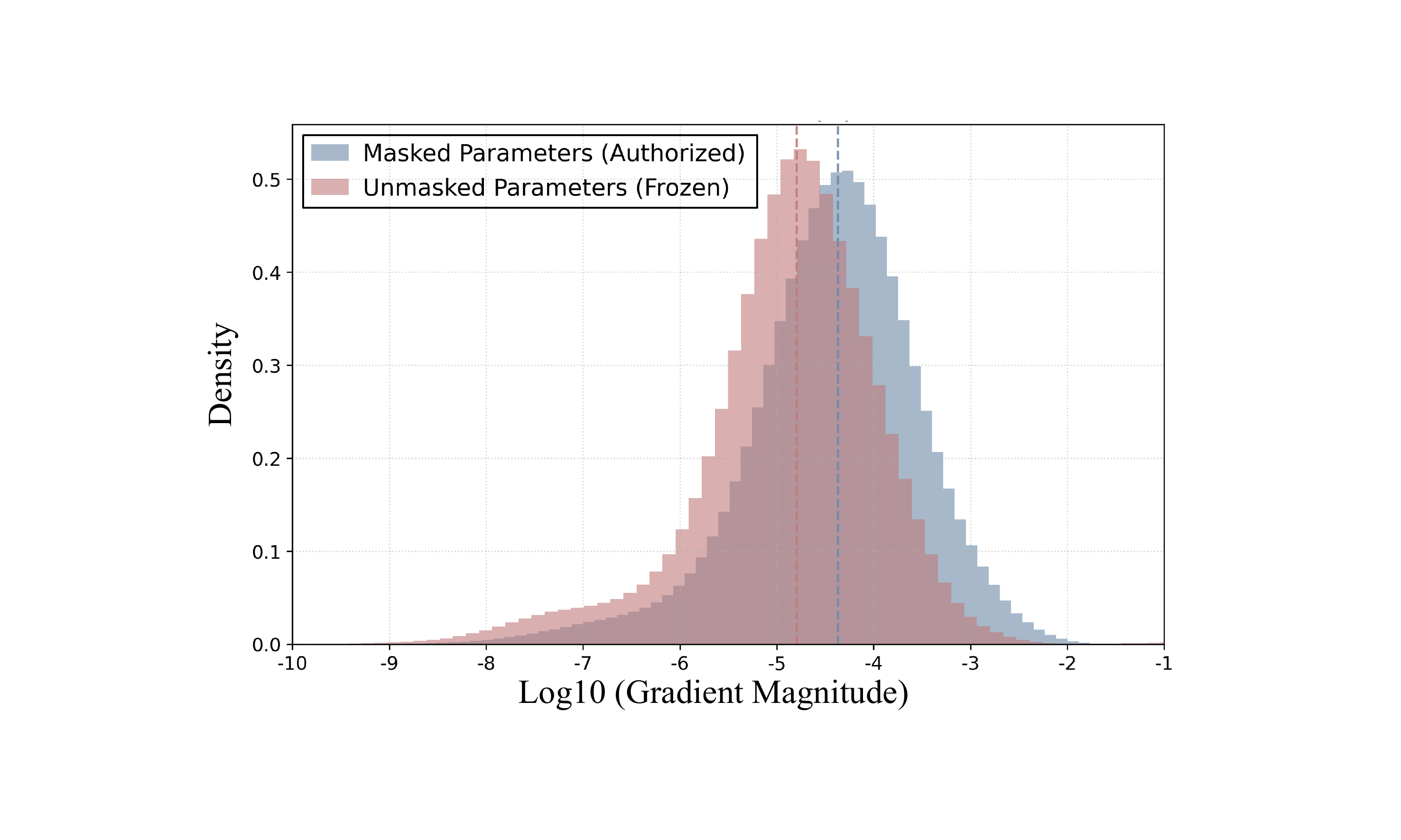}
    \caption{
    \textbf{Gradient magnitude distributions of masked (authorized) and unmasked (frozen) parameters on a held-out input.}
    For visualization clarity, gradients close to zero are omitted; these near-zero gradients predominantly correspond to unmasked (frozen) parameters.
    }
    \label{fig:gradient_dist}
\end{figure}

\subsection{Robustness Against Potential Mask Detection}

We consider a gradient-observing adversary who does not know the mask realization, the mask ratio, or the gradient magnitude distributions induced by masking.
In this setting, gradient magnitudes alone do not provide an identifiable signal for distinguishing masked and unmasked parameters from a single or limited number of observations, as absolute gradient scales lack semantic meaning without distributional priors.
Figure~\ref{fig:gradient_dist} shows the gradient magnitude distributions of masked and unmasked parameters for illustration.
Although a distributional shift can be observed \emph{a posteriori} when mask labels are known, the substantial overlap between the two distributions prevents reliable parameter-wise mask detection.
Consequently, without prior knowledge of the mask distribution, an adversary cannot effectively detect or exploit the private mask.

\section{Related Work}
\paragraph{Restriction on Foundation Model Fine-Tuning.}
Existing efforts on restricting foundation model fine-tuning primarily focus on preventing unauthorized adaptation to specific harmful datasets or domains. \citet{wang2024moderator} propose \emph{Moderator}, which performs an active defense by first specifying a target domain and then applying task-vector-based unlearning to suppress adaptation within that domain. \citet{deng2024sophon} introduce \emph{Sophon}, which mimics the fine-tuning dynamics of a target domain and explicitly constrains fine-tuning gradients during adaptation. A related line of work, often referred to as model immunization~\cite{zheng2024imma,li2025towards,sarkermodel}, injects task-specific disturbing signals into pretrained models to impair fine-tunability for particular concepts or domains.

Despite their effectiveness in controlled settings, these approaches share a fundamental limitation: they rely on prior knowledge of the downstream tasks to be restricted, and often require direct access to the corresponding fine-tuning data or task objectives. Such assumptions are unrealistic in many real-world scenarios, where model trainers have no visibility into how released models will be adapted. In contrast, our work aims to restrict fine-tunability as a general property of foundation models, proposing a pretraining-level algorithm that requires no preliminary knowledge of downstream tasks, data distributions, or fine-tuning procedures.

\paragraph{Lottery Ticket and Masked Training.}
The lottery ticket hypothesis shows that over-parameterized neural networks often contain sparse subnetworks that can achieve performance comparable to the full model~\cite{frankle2018lottery}. Subsequent work demonstrates that exploiting such lottery tickets can significantly improve training efficiency and stability~\cite{frankle2019stabilizing,morcos2019one}, with particular benefits in distributed learning systems~\cite{li2020lotteryfl} and large foundation models~\cite{panda2024lottery,yuan2025ks}. Most of these approaches identify lottery tickets after full or partial training. More recent work explores discovering lottery tickets at early stages of training to further reduce memory and computational overhead~\cite{chen2021earlybert}. 
Overall, existing studies on lottery tickets and masked training primarily aim to improve computational efficiency and scalability of machine learning systems. To the best of our knowledge, none of these works examine lottery tickets from the perspective of restricting model fine-tunability or consider their implications for security and control in foundation model deployment.

\section{Conclusion}

This paper studied the problem of building non-fine-tunable FMs, motivated by the need to reconcile open model release with control over downstream adaptation.
To this end, we proposed PMP, a pre-training paradigm that concentrates learning into a private sparse subnetwork and releases only the final dense parameters.
By withholding the mask, PMP prevents unauthorized fine-tuning from aligning with the optimization geometry induced during pre-training.
Our theoretical analysis shows that fine-tuning without access to the private mask optimizes a mismatched objective, injecting destabilizing updates along unadapted parameter directions.
Experiments on LLMs confirm this analysis: PMP preserves strong base performance while consistently degrading unauthorized fine-tuning across architectures, tasks, and hyperparameter choices.
Overall, PMP reframes non-fine-tunability as a pre-training-level property rather than a task-specific unlearning.
We hope this work encourages further research into controllable FM deployment, where openness and safety can coexist without requiring assumptions about downstream data or training procedures.

\section*{Impact Statement}

This paper presents work whose goal is to advance the field of machine learning by studying controllable pre-training mechanisms for foundation models. Our method aims to enable open model release while limiting unauthorized fine-tuning, which may help model developers better manage economic and safety risks associated with unrestricted downstream adaptation. We do not foresee immediate negative societal impacts beyond those commonly associated with the deployment of large-scale machine learning models, and we believe the ethical considerations raised by this work are largely aligned with existing discussions in the field.

\nocite{langley00}

\bibliography{example_paper}

@inproceedings{langley00,
 author    = {P. Langley},
 title     = {Crafting Papers on Machine Learning},
 year      = {2000},
 pages     = {1207--1216},
 editor    = {Pat Langley},
 booktitle     = {Proceedings of the 17th International Conference
              on Machine Learning (ICML 2000)},
 address   = {Stanford, CA},
 publisher = {Morgan Kaufmann}
}

@inproceedings{wang2024moderator,
  title={Moderator: Moderating text-to-image diffusion models through fine-grained context-based policies},
  author={Wang, Peiran and Li, Qiyu and Yu, Longxuan and Wang, Ziyao and Li, Ang and Jin, Haojian},
  booktitle={Proceedings of the 2024 on ACM SIGSAC Conference on Computer and Communications Security},
  pages={1181--1195},
  year={2024}
}

@inproceedings{deng2024sophon,
  title={Sophon: Non-fine-tunable learning to restrain task transferability for pre-trained models},
  author={Deng, Jiangyi and Pang, Shengyuan and Chen, Yanjiao and Xia, Liangming and Bai, Yijie and Weng, Haiqin and Xu, Wenyuan},
  booktitle={2024 IEEE Symposium on Security and Privacy (SP)},
  pages={2553--2571},
  year={2024},
  organization={IEEE}
}

@inproceedings{zheng2024imma,
  title={Imma: Immunizing text-to-image models against malicious adaptation},
  author={Zheng, Amber Yijia and Yeh, Raymond A},
  booktitle={European Conference on Computer Vision},
  pages={458--475},
  year={2024},
  organization={Springer}
}

@article{li2025towards,
  title={Towards resilient safety-driven unlearning for diffusion models against downstream fine-tuning},
  author={Li, Boheng and Gu, Renjie and Wang, Junjie and Qi, Leyi and Li, Yiming and Wang, Run and Qin, Zhan and Zhang, Tianwei},
  journal={arXiv preprint arXiv:2507.16302},
  year={2025}
}

@inproceedings{sarkermodel,
  title={Model Immunization by Trapping Harmful Finetuning},
  author={Sarker, Najibul Haque and Hakim, Zaber Ibn Abdul and Ishmam, Alvi Md and Tang, Chia-Wei and Thomas, Chris},
  booktitle={Lock-LLM Workshop: Prevent Unauthorized Knowledge Use from Large Language Models}
}

@article{frankle2018lottery,
  title={The lottery ticket hypothesis: Finding sparse, trainable neural networks},
  author={Frankle, Jonathan and Carbin, Michael},
  journal={arXiv preprint arXiv:1803.03635},
  year={2018}
}

@article{frankle2019stabilizing,
  title={Stabilizing the lottery ticket hypothesis},
  author={Frankle, Jonathan and Dziugaite, Gintare Karolina and Roy, Daniel M and Carbin, Michael},
  journal={arXiv preprint arXiv:1903.01611},
  year={2019}
}

@article{morcos2019one,
  title={One ticket to win them all: generalizing lottery ticket initializations across datasets and optimizers},
  author={Morcos, Ari and Yu, Haonan and Paganini, Michela and Tian, Yuandong},
  journal={Advances in neural information processing systems},
  volume={32},
  year={2019}
}

@article{li2020lotteryfl,
  title={Lotteryfl: Personalized and communication-efficient federated learning with lottery ticket hypothesis on non-iid datasets},
  author={Li, Ang and Sun, Jingwei and Wang, Binghui and Duan, Lin and Li, Sicheng and Chen, Yiran and Li, Hai},
  journal={arXiv preprint arXiv:2008.03371},
  year={2020}
}

@inproceedings{yuan2025ks,
  title={KS-Lottery: Finding Certified Lottery Tickets for Multilingual Transfer in Large Language Models},
  author={Yuan, Fei and Ma, Chang and Yuan, Shuai and Sun, Qiushi and Li, Lei},
  booktitle={Proceedings of the 2025 Conference of the Nations of the Americas Chapter of the Association for Computational Linguistics: Human Language Technologies (Volume 1: Long Papers)},
  pages={9077--9090},
  year={2025}
}

@article{panda2024lottery,
  title={Lottery ticket adaptation: Mitigating destructive interference in llms},
  author={Panda, Ashwinee and Isik, Berivan and Qi, Xiangyu and Koyejo, Sanmi and Weissman, Tsachy and Mittal, Prateek},
  journal={arXiv preprint arXiv:2406.16797},
  year={2024}
}

@inproceedings{chen2021earlybert,
  title={Earlybert: Efficient bert training via early-bird lottery tickets},
  author={Chen, Xiaohan and Cheng, Yu and Wang, Shuohang and Gan, Zhe and Wang, Zhangyang and Liu, Jingjing},
  booktitle={Proceedings of the 59th Annual Meeting of the Association for Computational Linguistics and the 11th International Joint Conference on Natural Language Processing (Volume 1: Long Papers)},
  pages={2195--2207},
  year={2021}
}

@article{guo2025deepseek,
  title={Deepseek-r1: Incentivizing reasoning capability in llms via reinforcement learning},
  author={Guo, Daya and Yang, Dejian and Zhang, Haowei and Song, Junxiao and Zhang, Ruoyu and Xu, Runxin and Zhu, Qihao and Ma, Shirong and Wang, Peiyi and Bi, Xiao and others},
  journal={arXiv preprint arXiv:2501.12948},
  year={2025}
}

@article{achiam2023gpt,
  title={Gpt-4 technical report},
  author={Achiam, Josh and Adler, Steven and Agarwal, Sandhini and Ahmad, Lama and Akkaya, Ilge and Aleman, Florencia Leoni and Almeida, Diogo and Altenschmidt, Janko and Altman, Sam and Anadkat, Shyamal and others},
  journal={arXiv preprint arXiv:2303.08774},
  year={2023}
}

@article{team2024gemini,
  title={Gemini 1.5: Unlocking multimodal understanding across millions of tokens of context},
  author={Team, Gemini and Georgiev, Petko and Lei, Ving Ian and Burnell, Ryan and Bai, Libin and Gulati, Anmol and Tanzer, Garrett and Vincent, Damien and Pan, Zhufeng and Wang, Shibo and others},
  journal={arXiv preprint arXiv:2403.05530},
  year={2024}
}

@article{comanici2025gemini,
  title={Gemini 2.5: Pushing the frontier with advanced reasoning, multimodality, long context, and next generation agentic capabilities},
  author={Comanici, Gheorghe and Bieber, Eric and Schaekermann, Mike and Pasupat, Ice and Sachdeva, Noveen and Dhillon, Inderjit and Blistein, Marcel and Ram, Ori and Zhang, Dan and Rosen, Evan and others},
  journal={arXiv preprint arXiv:2507.06261},
  year={2025}
}

@article{jaech2024openai,
  title={Openai o1 system card},
  author={Jaech, Aaron and Kalai, Adam and Lerer, Adam and Richardson, Adam and El-Kishky, Ahmed and Low, Aiden and Helyar, Alec and Madry, Aleksander and Beutel, Alex and Carney, Alex and others},
  journal={arXiv preprint arXiv:2412.16720},
  year={2024}
}

@article{touvron2023llama,
  title={Llama: Open and efficient foundation language models},
  author={Touvron, Hugo and Lavril, Thibaut and Izacard, Gautier and Martinet, Xavier and Lachaux, Marie-Anne and Lacroix, Timoth{\'e}e and Rozi{\`e}re, Baptiste and Goyal, Naman and Hambro, Eric and Azhar, Faisal and others},
  journal={arXiv preprint arXiv:2302.13971},
  year={2023}
}

@article{lin2024awq,
  title={Awq: Activation-aware weight quantization for on-device llm compression and acceleration},
  author={Lin, Ji and Tang, Jiaming and Tang, Haotian and Yang, Shang and Chen, Wei-Ming and Wang, Wei-Chen and Xiao, Guangxuan and Dang, Xingyu and Gan, Chuang and Han, Song},
  journal={Proceedings of machine learning and systems},
  volume={6},
  pages={87--100},
  year={2024}
}

@article{singh2024rethinking,
  title={Rethinking interpretability in the era of large language models},
  author={Singh, Chandan and Inala, Jeevana Priya and Galley, Michel and Caruana, Rich and Gao, Jianfeng},
  journal={arXiv preprint arXiv:2402.01761},
  year={2024}
}

@article{hu2022lora,
  title={Lora: Low-rank adaptation of large language models.},
  author={Hu, Edward J and Shen, Yelong and Wallis, Phillip and Allen-Zhu, Zeyuan and Li, Yuanzhi and Wang, Shean and Wang, Lu and Chen, Weizhu and others},
  journal={ICLR},
  volume={1},
  number={2},
  pages={3},
  year={2022}
}

@article{yu2025dapo,
  title={Dapo: An open-source llm reinforcement learning system at scale},
  author={Yu, Qiying and Zhang, Zheng and Zhu, Ruofei and Yuan, Yufeng and Zuo, Xiaochen and Yue, Yu and Dai, Weinan and Fan, Tiantian and Liu, Gaohong and Liu, Lingjun and others},
  journal={arXiv preprint arXiv:2503.14476},
  year={2025}
}

@article{li2026generative,
  title={When Generative AI Is Intimate, Sexy, and Violent: Examining Not-Safe-For-Work (NSFW) Chatbots on FlowGPT},
  author={Li, Xian and Han, Yuanning and Liu, Di and An, Pengcheng and Niu, Shuo},
  journal={arXiv preprint arXiv:2601.14324},
  year={2026}
}

@inproceedings{poppi2024safe,
  title={Safe-clip: Removing nsfw concepts from vision-and-language models},
  author={Poppi, Samuele and Poppi, Tobia and Cocchi, Federico and Cornia, Marcella and Baraldi, Lorenzo and Cucchiara, Rita},
  booktitle={European Conference on Computer Vision},
  pages={340--356},
  year={2024},
  organization={Springer}
}

@inproceedings{wang2018glue,
  title={GLUE: A multi-task benchmark and analysis platform for natural language understanding},
  author={Wang, Alex and Singh, Amanpreet and Michael, Julian and Hill, Felix and Levy, Omer and Bowman, Samuel},
  booktitle={Proceedings of the 2018 EMNLP workshop BlackboxNLP: Analyzing and interpreting neural networks for NLP},
  pages={353--355},
  year={2018}
}
\bibliographystyle{icml2025}

\newpage
\appendix
\onecolumn
\section{Additional Experimental Set-Up}

\paragraph{Hardware and Hyperparameters.}
All pre-training runs are executed on a single NVIDIA H20 GPU. We optimize using AdamW with a cosine learning-rate schedule, using a base learning rate of $\eta=2\times10^{-5}$, 2{,}000 learning-rate warmup optimizer updates, and a total of 20{,}000 optimizer updates. We use a micro-batch size of 4 sequences with gradient accumulation of 8, resulting in an effective batch size of 32 sequences per optimizer update. We clip the global gradient norm to 1.0 and fix the random seed to 42.

\end{document}